\documentclass{article}
\usepackage{ijcai26}

\usepackage{times}
\usepackage{soul}
\usepackage{url}
\usepackage[hidelinks]{hyperref}
\usepackage[utf8]{inputenc}
\usepackage[small]{caption}
\usepackage{graphicx}
\usepackage{amsmath}
\usepackage{amsthm}
\usepackage{booktabs}
\usepackage{algorithm}
\usepackage{algorithmic}
\usepackage[switch]{lineno}

\usepackage{enumitem}

\usepackage{pifont}       
\usepackage{makecell}     
\usepackage{underscore}
\usepackage{mathrsfs}

\title{CI4A: Semantic Component Interfaces for Agents Empowering Web Automation}

\author{
Zhi Qiu$^1$\and
Jiazheng Sun$^{2,1}$\and
Chenxiao Xia$^{1}$\and
Jun Zheng$^1$\And
Xin Peng$^2$\\
\affiliations
$^1$ School of Cyberspace Science and Technology, Beijing Institute of Technology\\
$^2$ College of Computer Science and Artificial Intelligence, Fudan University\\
\emails
zhengjun@bit.edu.cn,
pengxin@fudan.edu.cn
}

\begin{document}

\maketitle

\begin{abstract}
    While Large Language Models demonstrate remarkable proficiency in high-level semantic planning, they remain limited in handling fine-grained, low-level web component manipulations. To address this limitation, extensive research has focused on enhancing model grounding capabilities through techniques such as Reinforcement Learning. However, rather than compelling agents to adapt to human-centric interfaces, we propose constructing interaction interfaces specifically optimized for agents. This paper introduces Component Interface for Agent (CI4A), a semantic encapsulation mechanism that abstracts the complex interaction logic of UI components into a set of unified tool primitives accessible to agents. We implemented CI4A within Ant Design, an industrial-grade front-end framework, covering 23 categories of commonly used UI components. Furthermore, we developed a hybrid agent featuring an action space that dynamically updates according to the page state, enabling flexible invocation of available CI4A tools. Leveraging the CI4A-integrated Ant Design, we refactored and upgraded the WebArena benchmark to evaluate existing SoTA methods. Experimental results demonstrate that the CI4A-based agent significantly outperforms existing approaches, achieving a new SoTA task success rate of 86.3\%, alongside substantial improvements in execution efficiency.
\end{abstract}

\section{Introduction}
With the evolution of Large Language Models, Web Agents have demonstrated immense potential for complex, open-ended tasks \cite{yang2025agentic,ning2025survey,zhang2025large}. While they excel in information-intensive scenarios with heavy reasoning but light interaction like Deep Research \cite{tao2025webshaper,yu2026webanchor}, they struggle with operation-intensive tasks requiring precise UI manipulation and multi-step state changes. Constrained by limited precision and robustness at the interaction level, their task success rates remain significantly below the standards required for industrial applications.

The fundamental cause of this disparity lies in the fact that modern web pages are constructed for human users capable of processing continuous visual feedback and executing millisecond-level motor control, whereas agents are constrained by a discrete, high-latency Observation-Reasoning-Action loop. On the information side, web interfaces universally adopt human-centric Progressive Disclosure mechanisms (e.g., \textit{collapsible panels}). This forces agents to expend substantial reasoning resources on tedious exploratory actions merely to complete the decision-making context \cite{gu2024your,koh2024tree,zhang2025webpilot}. On the execution side, interaction modalities centered on visual coordinates and low-level events confine agents to a low-level action space, compelling them to decompose high-dimensional user intents into lengthy chains of atomic operations \cite{erdogan2025plan,he2025branch}. Given the agent's discontinuous operational mode, such long sequences not only significantly reduce execution efficiency but also make the process highly susceptible to error accumulation from minor low-level missteps,  leading to the collapse of the overall high-level task.

\begin{figure*}[t]
\centerline{\includegraphics[width=0.8\textwidth]{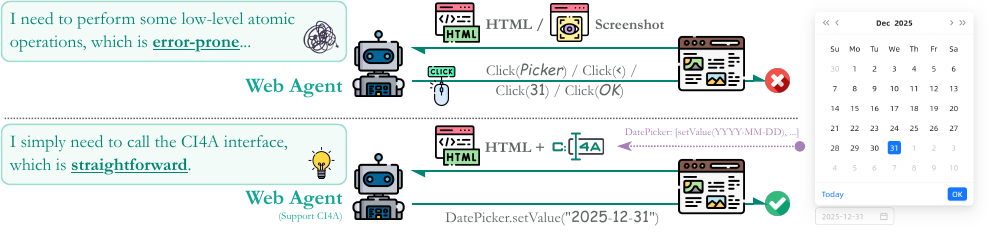}}
\caption{\textbf{Fragile Atomic Operations vs. Robust CI4A.} In contrast to traditional methods entangled in error-prone multi-step chains for date selection, CI4A collapses complex logic into \textbf{deterministic single-step invocations}, eliminating the uncertainty of long-sequence execution.}
\label{fig1}
\end{figure*} 

To address this challenge, existing research primarily focuses on two directions: first, enhancing internal capabilities by leveraging supervised fine-tuning \cite{liu2023webglm,gur2023real,lee2025learning} or Reinforcement Learning \cite{liu2025infigui,zhang2025prune4web,chen2026gui} to improve the agent's grounding precision; second, constructing High-level Skills, which attempts to encapsulate complex low-level interaction sequences into callable tools (e.g., search, sort) to simplify the operation space \cite{yang2025ultracua,pang2025assimilation,he2025recon}. However, the former fails to eliminate the dependency on the fragile DOM, while the latter often requires complex reverse engineering \cite{prabhu2025walt} or heuristic rules to synthesize skills \cite{wang2024agent}, making generality difficult to ensure.

Rather than patching or applying secondary abstractions to human-centric interfaces, we propose reconstructing the interaction paradigm from the source to achieve native alignment between interface semantics and agent reasoning logic. Modern web applications are universally based on modular UI component architectures, which naturally encapsulate high-level functional semantics. This offers an ideal entry point to transcend low-level DOM or GUI limitations and reshape a component-centric interaction paradigm.

To this end, we introduces Component Interface for Agent (CI4A), a semantic encapsulation mechanism designed to bridge the gap between high-level task intents and low-level web interactions. CI4A abstracts complex UI components into a set of unified, standardized \textbf{Tool Primitives} optimized for agent invocation patterns. By exposing structured interfaces for state retrieval and functional operations, CI4A allows agents to bypass the noisy and fragile low-level DOM, enabling efficient interaction directly at the component level with explicit semantics.

Building upon CI4A, we developed Eous, a hybrid-architecture agent capable of dynamically extending its action space based on the current page state to include and utilize tools exposed by CI4A components. We instantiated CI4A within the industrial-grade Ant Design UI framework, covering 23 categories of commonly used UI components, and utilized this implementation to refactor the Playground scenarios of the WebArena dataset for evaluation. Experimental results indicate that the CI4A-based Eous achieved a task success rate of 86.3\% across 182 tasks. While significantly reducing the number of interaction steps, it substantially outperforms existing State-of-The-Art (SoTA) methods (70.3\%). The contributions of this paper are summarized as follows.

\begin{itemize}[left=0pt, nosep] 
    \item We propose \textbf{CI4A}, a semantic encapsulation protocol that transforms UI components into standardized tool primitives, bridging the gap between interface semantics and agent reasoning. We provide \textbf{AntDX}, a reference implementation covering 23 core Ant Design components to validate its generality. 
    
    \item We design \textbf{Eous}, an agent featuring a hybrid action space that dynamically combines high-level semantic tools with low-level operation fallbacks. This layered strategy enables atomic execution of high-dimensional intents, significantly enhancing both effectiveness and efficiency. 
    
    \item We evaluate our approach on a semantically reconstructed WebArena benchmark. Experiments show that \textbf{CI4A enables Eous to significantly outperform existing SoTA methods}.
\end{itemize}

Our work is available at \href{https://anonymous.4open.science/r/CI4A-74EC}{Anonymous GitHub Link}.

\section{Related Work}
\noindent \textbf{Perception and Grounding in Web Agents.} Existing perception paradigms fall primarily into two categories. Structure-based methods (e.g., WebArenaBase \cite{zhou2023webarena}, SteP \cite{sodhi2024step}) leverage the Accessibility Tree or HTML for perception, whereas visual-based methods utilize screenshots to achieve grounding via coordinate regression (SeeClick \cite{cheng2024seeclick}) or visual prompting (WebVoyager \cite{he2024webvoyager} combined with Set-of-Mark). Tools such as OmniParser \cite{wan2024omniparser} have also emerged to facilitate screen parsing. However, these approaches remain constrained by the Progressive Disclosure mechanisms inherent in the modern Web. Agents are unable to directly perceive implicit states, such as lazy-loaded content or collapsed panels, compelling them to rely on inefficient physical exploration. Furthermore, visual methods \cite{qin2025ui} continue to struggle with robustness bottlenecks, including hallucinations and coordinate drift.

\noindent \textbf{Action Spaces and Skill Acquisition.} The granularity of action abstraction determines interaction efficiency. Mainstream approaches relying on atomic operations (e.g., click, type, scroll) \cite{yang2025agentoccam} suffer from lengthy decision-making chains that are prone to error accumulation. To enhance efficiency, systems like SkillWeaver \cite{zheng2025skillweaver} and ASI \cite{wang2025inducing} attempt to extract high-level skills, while WALT \cite{prabhu2025walt} synthesizes callable tools by mining APIs. However, such post-hoc abstractions incur high construction costs and lack generalizability. Moreover, they prove highly fragile in the face of minor modifications to frontend layouts or APIs.

\noindent \textbf{Paradigms of Agent-Web Interaction.} Neither traditional UIs (tailored for humans) nor Web APIs (designed for developers) are optimized for the decision-making characteristics of agents. Consequently, \cite{lu2025build} proposed the Agentic Web Interface (AWI) concept, resonating with the semantic enhancement philosophy of WAI-ARIA \cite{craig2009accessible}. Although AWI delineates the necessary transition from adapting to human interfaces to constructing agent-centric interfaces, there remains a significant gap. Specifically, there is currently no concrete protocol implementation that integrates seamlessly with the modern component-based frontend ecosystem while maintaining low intrusiveness.

\section{Methodology}

\subsection{The CI4A Protocol}
The objective of CI4A is to establish a standard Web interaction protocol tailored for agents. This protocol aims to bridge the semantic gap between human-computer interfaces and agent cognition. By defining the semantic interfaces that UI components expose to agents and standardizing the interaction patterns for environment perception and action execution, CI4A transforms Web interaction from fragile reverse engineering based on rendered results into a robust paradigm of forward invocation based on component instances.

\subsubsection{UI Component Interface}
Within the CI4A protocol, we abstract the interaction capabilities exposed by any UI component $C$ into a semantic triplet $\langle \mathcal{S}, \mathcal{T}, \mathcal{M} \rangle$. This definition constitutes a static contract governing the communication between the agent and the component:

\begin{figure}[htbp]
\centering
\includegraphics[width=\linewidth]{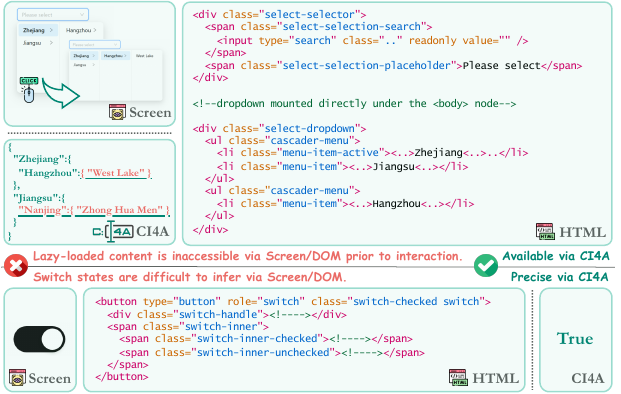}
\caption{CI4A’s Semantic State View ensures accurate state perception for hidden lazy-loaded data and ambiguous DOM states.}

\label{fig2}
\end{figure}

\noindent \textbf{Semantic State View $\mathcal{S}$} acts as a direct mapping of the component's internal data model, enabling the agent to bypass the rendering layer and access pure business data directly from memory (e.g., Vue's reactive objects or React's State/Props). This mechanism reduces the Context Encoding Cost from $O(N)$, which scales with DOM tree depth, to $O(1)$, which is decoupled from rendering details. Consequently, it fundamentally enhances the completeness and fidelity of perception. Specifically, it enables the agent to penetrate Progressive Disclosure mechanisms, such as lazy loading, to acquire a complete view of the data while accessing raw business values. This effectively eliminates the noise and errors typically introduced by DOM parsing and CV recognition.

\noindent  \textbf{Executable Toolset $\mathcal{T}$} defines a collection of Atomic Operation Primitives that encapsulate the logic for component state mutation. Whether derived from the component's native public methods or the direct manipulation of internal states, $\mathcal{T}$ uniformly abstracts these operations as standard function calls at the protocol level. By collapsing error-prone, low-level multi-step operation chains into single atomic invocations, this mechanism mitigates hallucinations and cascading errors arising from long-sequence execution. Furthermore, $\mathcal{T}$ establishes an Optimal Granularity between low-level DOM events and high-level business logic. This abstraction shields the agent from tedious execution details while ensuring the tools remain generalizable and reusable across scenarios.
    
\noindent  \textbf{Interaction Metadata $\mathcal{M}$} defines a Structured Contract for tool invocation, explicitly describing parameter data types, enumeration ranges, and mandatory constraints based on component attributes. This mechanism addresses the issue of constraint invisibility, where restrictions are imperceptible solely through visual or DOM cues. Such invisibility often leads to execution failures because Large Models lack prior knowledge of input formats. By dynamically transforming $\mathcal{M}$ into Schema definitions within the Prompt, CI4A effectively bridges the agent's knowledge gap of business rules. This enables the agent to generate legitimate actions based on explicit parametric knowledge, thereby preventing blind trial-and-error driven by information insufficiency.

\subsubsection{Agent-UI Interaction Patterns}
In the CI4A protocol, we standardize the entire process by which the agent utilizes the aforementioned interfaces to perceive the environment and execute actions into a set of \textbf{canonical dynamic interaction patterns}. This definition establishes the dynamic mechanism governing the collaboration between the agent and the runtime Web UI environment:

\begin{figure}[htbp]
\centering
\includegraphics[width=\linewidth]{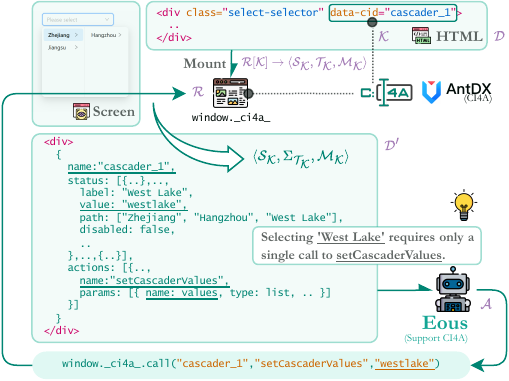}
\caption{\textbf{CI4A workflow.} Components register with a global registry, allowing agents to retrieve the triplet $\langle \mathcal{S_\mathcal{K}}, \Sigma_{\mathcal{T_\mathcal{K}}}, \mathcal{M_\mathcal{K}} \rangle$ for hybrid view construction. The agent then derives tools invocations and parameters from this view and executes tools via the registry.}
\label{fig3}
\end{figure}

\noindent \textbf{Hybrid View Construction.} To balance the efficient perception of spatial Layout Structure and component semantics, CI4A proposes a hybrid view construction strategy that transforms the raw DOM tree $\mathcal{D}$ into a semantic hybrid tree $\mathcal{D'}$. During the component rendering stage, UI component libraries implementing CI4A must assign a unique identifier $\mathcal{K}$ to each component instance and explicitly annotate it on the corresponding DOM root node. This process remains transparent to existing Web development workflows and does not interfere with standard DOM debugging or parsing. For an agent $\mathcal{A}$ supporting CI4A, the strategy mandates specific behaviors when traversing $\mathcal{D}$. For unmarked structural nodes (such as div or span tags responsible for layout), $\mathcal{A}$ retains their hierarchical relationships to preserve spatial context. Conversely, for nodes annotated with $\mathcal{K}$, $\mathcal{A}$ replaces the underlying DOM with the component's corresponding semantic description node, which comprises the semantic state view $\mathcal{S}$, the function signatures $\Sigma_{\mathcal{T}}$ of the executable tool set, and the interaction metadata $\mathcal{M}$.

\noindent \textbf{Global Registration \& Indexing.} To enable the agent to invoke tools $\mathcal{T}$ declared within the view, CI4A mandates that the runtime environment maintain a global registry $\mathcal{R}$. This registry $\mathcal{R}$ must be exposed within the browser's JavaScript runtime global object (e.g., window), serving as a conduit for communication between the agent and component instances. Upon component mounting, the instance $\mathcal{K}$ automatically establishes the mapping $\mathcal{R}[\mathcal{K}] \rightarrow \langle \mathcal{S_\mathcal{K}}, \mathcal{T_\mathcal{K}}, \mathcal{M_\mathcal{K}} \rangle$, which is subsequently removed upon unmounting. The agent $\mathcal{A}$ can access $\mathcal{R}$ via the global object to directly index component instances using the key $\mathcal{K}$, thereby retrieving $\langle \mathcal{S_\mathcal{K}}, \Sigma_{\mathcal{T_\mathcal{K}}}, \mathcal{M_\mathcal{K}} \rangle$ or accessing the execution handle $t \in \mathcal{T}_\mathcal{K}$.

\noindent \textbf{Deterministic Execution.} To achieve determinism and observability in the interaction process, CI4A requires the agent $\mathcal{A}$ to synthesize action commands $t$ based primarily on the action definitions provided by $\Sigma_{\mathcal{T}}$ and the constraints defined in $\mathcal{M}$, subsequently issuing an invocation request to $\mathcal{R}$. Simultaneously, CI4A mandates that $\mathcal{R}$ perform runtime parameter validation prior to instruction execution. If a parameter violation is detected, the component is disabled, or $\mathcal{K}$ is invalid, $\mathcal{R}$ must terminate execution and return a structured exception. This mechanism facilitates the construction of a rapid feedback loop devoid of visual model intervention. It effectively resolves the confirmation dilemma faced by traditional visual agents, where determining whether an action was effective remains difficult even if the correct element was clicked.

\subsection{AntDX: A Reference Implementation of CI4A Based on the Ant Design UI Library}
\label{AntDX}
Ant Design stands as one of the de facto standards for enterprise Web development. To validate the completeness of the CI4A protocol, we developed a reference implementation named \textbf{AntDX}. By applying \textbf{Intrusive Instrumentation} to the source code of 23 core components, AntDX assumes control over the component lifecycle and event streams, achieving a deep integration of the protocol logic.

\subsubsection{Core Instrumentation Components}

 \textbf{Global Registrar.} Exposed via \textit{\texttt{window.__ci4a__}}, this component serves as the unified gateway for interactions between the Agent and UI components. It dynamically maintains the Transceivers for all active components and routes requests to provide two core capabilities: \textbf{a) State Provisioning ($get\mathcal{S}tatus()$):} Supplies the business semantic state $\mathcal{S}$, tool signatures $\Sigma_{\mathcal{T}}$, and metadata $\mathcal{M}$. \textbf{b) Action Execution ($call\mathcal{T}ool()$):} Triggers deterministic business logic within the component.

\noindent\textbf{Component Transceiver.} An execution unit embedded within the component, the Transceiver establishes a semantic bridge between the low-level instance and the Agent. It comprises three distinct modules:
 \begin{itemize}[left=0pt, nosep]
    \item \textbf{Auto-Register:} Automatically manages the mapping between the instance and the Global Registrar during the Mount/Unmount lifecycle stages. Additionally, it binds a unique identifier to the data-cid attribute of the DOM.
    \item \textbf{Props Listener:} Reactively monitors component Props. Upon receiving a request for the business semantic state, it filters out noisy fields based on a whitelist, subsequently generating and returning a pure semantic view that encapsulates the core business state.
    \item \textbf{Dispatcher:} Monitors instructions from the Global Registrar and orchestrates their dispatch. Semantic state requests are routed to the Props Listener. Action execution requests are prioritized for forwarding to standard event handlers (e.g., \texttt{\textit{update:value}}), allowing the component to be driven without disrupting the normal event flow. In cases where an explicit handler for the action is absent, the Dispatcher adopts a fallback strategy of directly manipulating the component's internal state.
\end{itemize}

\subsubsection{Implementation of UI Component Interfaces}
Based on the distinct roles components play in Agent decision-making and execution during Web tasks, we categorize them into three types: Navigation (e.g., Menu), Data Entry (e.g., DatePicker), and Data Display (e.g., Table). For each category, we implemented a mapping from the raw UI to the semantic triplet $\langle \mathcal{S}, \mathcal{T}, \mathcal{M} \rangle$.

\noindent \textbf{Semantic State View $\mathcal{S}$.} Navigation components expose reachable routes or step nodes. Data Entry components provide field identifiers, input guidance, and the set of currently selected or input values. Data Display components present structured data content within the viewport (e.g., table headers and row data).

\noindent \textbf{Executable Tool Set $\mathcal{T}$.} Navigation components provide \textit{\texttt{navigateTo}} to implement direct transitions. Data Entry components offer \textit{\texttt{setValue}} and \textit{\texttt{submit}} to efficiently complete data entry and trigger validation. Data Display components provide \textit{\texttt{sort}} and \textit{\texttt{filter}} to adjust data presentation and precisely screen for required information.

\noindent  \textbf{Interaction Metadata $\mathcal{M}$.} This comprises metadata objects that are manually constructed and injected at compile time (exhaustively defining field types, etc.). For Data Entry components, $\mathcal{M}$ additionally includes input constraints calculated at runtime, such as date formats and numerical ranges.

\subsection{Eous: A Hybrid Action Space Agent}

To validate the effectiveness of the CI4A protocol, we developed Eous based on Fairy \cite{sun2025fairy}, a GUI agent framework featuring a typical Planner-Decider-Reflector closed loop. Benefiting from Fairy's highly decoupled software design pattern, we seamlessly migrated its originally mobile-oriented perception module to the Web environment. Furthermore, we refactored the Action Decider to introduce a Hybrid Action Space mechanism. This mechanism enables Eous to transcend the limitations of low-level atomic operations by dynamically invoking the component-level semantic tools defined in Section \ref{AntDX}. Specifically, Eous implements a semantic-atomic hybrid architecture within two critical phases: perception and decision-making.

In the perception phase, Eous introduces a modality-agnostic semantic injection mechanism that is compatible with and enhances mainstream structural and visual perception paradigms:

\begin{table*}
    \centering
    \begin{tabular*}{\linewidth}{@{\extracolsep{\fill}}lccccc}
        \toprule
        Framework     & Model     & Accessibility Tree & Screenshot & \makecell{Success Rate (\%)} & \makecell{Average Steps} \\
        \midrule
        WebarenaBase \cite{zhou2023webarena}     & GPT-5     & \ding{51}  & \ding{55}  & 26.4          & 10.0        \\
        AgentOccam (S) \cite{yang2025agentoccam} & GPT-5     & \ding{51}  & \ding{55}  & 69.2          & 9.4         \\
        AgentOccam (V) \cite{yang2025agentoccam} & GPT-5     & \ding{51}  & \ding{51}  & 70.3          & 10.4        \\
        MidScene \cite{Midscene}      & Seed-1.8  & \ding{55}  & \ding{51}  & 61.5          & 13.0        \\
        Eous (S)       & GPT-5     & \ding{51}  & \ding{55}  & \underline{\textbf{75.3}}  & \underline{\textbf{4.2}}     \\
        Eous (V)       & GPT-5     & \ding{55}  & \ding{51}  & \underline{\textbf{86.3}}  & \underline{\textbf{4.7}}     \\
        \bottomrule
    \end{tabular*}
    \caption{Main results on the WebArena benchmark (S: Structural Input, V: Visual Input). Eous(V) achieves a SoTA success rate of 86.3\% and reduces average steps to 4.7, significantly outperforming existing baselines.}
    \label{tab:framework_comparison}
\end{table*}

\textbf{For Structural Input,} The agent scans the DOM / Accessibility tree for instrumented nodes $\mathcal{K}$ bearing the data-cid attribute. It then retrieves the semantic triplet $\langle \mathcal{S}, \Sigma_{\mathcal{T}}, \mathcal{M} \rangle$ via \textit{\texttt{window.__ci4a__}} and directly replaces the original subtree. This semantic-structural hybrid view fundamentally eliminates redundant node noise while preserving the original page topology.

\textbf{For Visual Input,} Eous integrates Set-of-Mark (SoM) technology to overlay visual ID masks onto interactive elements. It injects the retrieved semantic triplet $\langle \mathcal{S}, \Sigma_{\mathcal{T}}, \mathcal{M} \rangle$ into the Prompt as supplementary context strongly correlated with the visual ID. In this configuration, $\mathcal{S}$ functions as a dynamic legend that elucidates implicit states. Meanwhile, $\Sigma_{\mathcal{T}}$ and $\mathcal{M}$ bind executable semantic operation contracts, allowing the agent to bypass low-level coordinate operations and directly plan high-level component actions.

In the decision-making phase, Eous's action space is defined as the union $a = a_{\Sigma_{\mathcal{T}}} \cup a_{atomic}$. We define a unified primitive, $call(k, \tau, p)$, where $k \in \mathcal{K}$ represents the target component, $\tau \in \Sigma_{\mathcal{T}_k}$ denotes the tool signature, and $p$ is the set of parameters that must satisfy the constraints of metadata $\mathcal{M}_\mathcal{K}$. Execution adheres to a Semantic First strategy. This strategy prioritizes the invocation of standardized tools $a_{\Sigma_{\mathcal{T}}}$ exposed by components (e.g., \textit{\texttt{navigateTo}}, \textit{\texttt{setValue}}). The agent adaptively falls back to low-level WebDriver atomic operations (e.g., \textit{\texttt{click}}, \textit{\texttt{type}}) only when applicable semantic tools are absent or execution is obstructed. This approach strikes a balance between interaction efficiency and generality.

\section{Experiment}

\subsection{Setup}
\textbf{Benchmark.} We selected WebArena as our primary benchmark to evaluate the CI4A protocol within a highly realistic and reproducible environment. However, validating a new interaction protocol presents a cold start dilemma far more severe than validating a new algorithm. Existing benchmarks, such as WebArena, are tailored for agents relying on standard DOM or visual interactions, and their underlying UI component libraries cannot be swapped directly. Consequently, we performed a semantic-preserving refactoring of the WebArena frontend using the standard Ant Design library. The refactored pages strictly maintain the original layout, content, and business logic, replacing native components solely with their Ant Design equivalents. While this ensures that visual and logical flows remain unchanged, the resulting DOM tree structure becomes significantly deeper and more complex due to component encapsulation. This design minimizes interference arising from environmental discrepancies, thereby ensuring a fair experimental comparison.

Building on this foundation, and leveraging the API consistency between AntDX and the standard library, we seamlessly integrated AntDX components via dependency replacement without modifying any business code. This implementation validates the low intrusiveness of the CI4A protocol and demonstrates that it imposes zero additional development costs on existing frontend engineering systems.

\noindent \textbf{Baseline.} We selected two groups of SoTA baselines for rigorous comparison against Eous:

\begin{itemize}[left=0pt, nosep]
    \item \textbf{Unimodal Structural Perception:} WebArenaBase and AgentOccam (text-only mode). Both are powered by GPT-5 and establish the performance benchmark for agents relying on the Accessibility Tree.
    \item \textbf{Multimodal Perception:} AgentOccam (standard mode, GPT-5) and the end-to-end framework MidScene (powered by native Seed-1.8 \cite{seedseed1}). These establish benchmarks for visually enhanced and end-to-end agents.
\end{itemize}

Correspondingly, we configured Eous (GPT-5) in two modes: structural input mode (Accessibility Tree infused with semantic triplets) and visual input mode (SoM combined with semantic triplet context supplementation).

All GPT-5 experiments were conducted via the OpenAI API (Context=128k). For all agents, the maximum number of interaction steps per task was set to 30; exceeding this limit constitutes a failure.

\noindent \textbf{Metrics.} We adopt a dual-metric evaluation system:

\begin{itemize}[left=0pt, nosep]
    \item \textbf{Success Rate(SR):} Measures effectiveness. Adhering to WebArena's Functional Correctness standard, this validates whether the final page or database state aligns with the Oracle's expectation upon task completion.
    \item \textbf{Average Steps:} Measures efficiency. By quantifying the decision turns, this metric evaluates the chain collapse effect of high-level semantic tools relative to low-level atomic operations.
\end{itemize}

\subsection{Main Result}

\begin{figure*}[t]
\centerline{\includegraphics[width=1\textwidth]{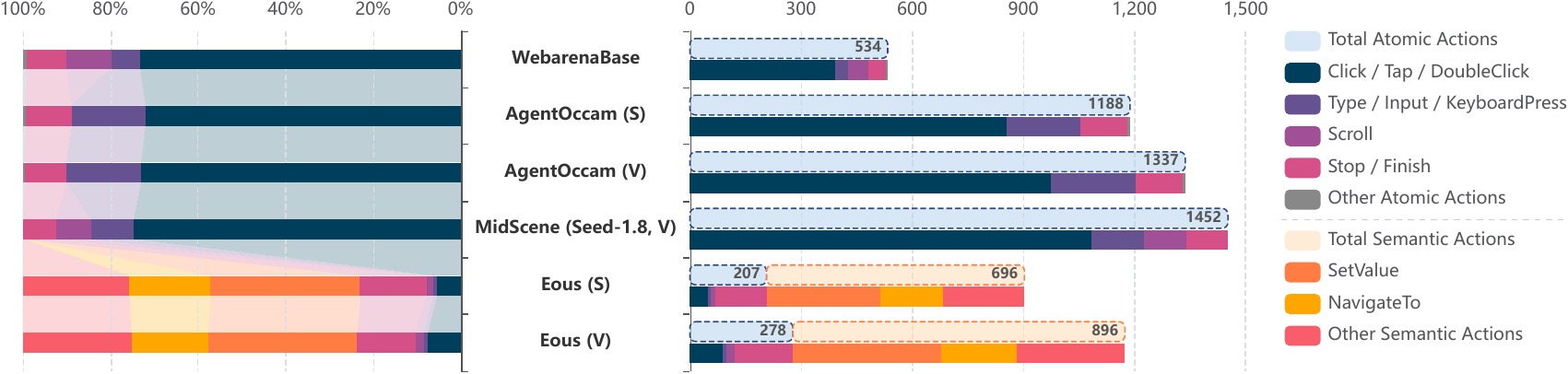}}
\caption{\textbf{Action space distribution of Eous vs. Baselines.} Compared to baselines that rely on lengthy low-level atomic actions, Eous leverages high-level semantic tools via CI4A to significantly collapse physical interaction chains, thereby reducing total steps.}
\label{fig4}
\end{figure*} 

We report the overall performance of Eous on the WebArena dataset in Table 1. Following the semantic-preserving refactoring using the standard Ant Design library, the WebArena environment exhibits an 34\% increase in Accessibility Tree depth compared to the original version (increased from 12.1 to 16.2 levels), thereby imposing more rigorous challenges on the agents under evaluation. Experimental results demonstrate that Eous exhibits significant performance advantages across all test dimensions. Whether compared to unimodal perception based on the Accessibility Tree or current state-of-the-art multimodal end-to-end models, Eous, empowered by the CI4A protocol, achieves a substantial improvement in SR while significantly reducing the length of interaction sequences.

Specifically, our experimental findings are as follows:

\textbf{$\triangleright$ Eous achieves SoTA task success rates across different modalities.} As shown in Table 1, the CI4A protocol significantly extends the operational boundaries of general LLMs within complex GUI environments: 1) \textit{In the unimodal structural perception track.} Eous (Structured Mode) with 75.3\% average success rate significantly surpassed WebArenaBase (26.4\%) and text-only mode AgentOccam (69.2\%). This indicates that compared to standard, lossy compressed Accessibility Tree, the semantic view provided by CI4A provides a better context foundation for the decision process, eliminating uncertainty at the planning level. 2) \textit{In the multimodal perception track.} Eous (Vision Mode) achieves an average success rate of 86.3\%. It not only surpasses the general-purpose AgentOccam (70.3\%) but also outperforms the specialized end-to-end model Seed-1.8 (61.5\%), notably without any domain-specific fine-tuning. This demonstrates that CI4A's high-level tool abstraction effectively circumvents the limitations of visual perception and execution in complex scenarios, achieving an execution fidelity akin to What You Intend Is What You Get.

\textbf{$\triangleright$ The hybrid action space achieves an order-of-magnitude improvement in decision-making efficiency.} As shown in Table 1, Eous drastically reduces the average decision sequence length required to complete tasks from a baseline of 10.7 steps to 4.5 steps, representing a relative reduction of 57.5\%. This confirms that the synergy between the high-level semantic tools $\Sigma_{\mathcal{T}}$ introduced by CI4A and the comprehensive semantic view $\mathcal{S}$ effectively truncates decision sequences. Consequently, this substantially accelerates task completion, given that action decision-making constitutes the primary time-consuming operation.

\subsection{In-depth Analysis}

\subsubsection{Data Insights}

To investigate the underlying mechanisms driving Eous's performance leap, we conducted an in-depth analysis across two dimensions: action space distribution and robustness against task complexity.

\textbf{$\triangleright$ The semantic escalation of interaction paradigms drives the collapse of physical sequences.} As illustrated in Figure \ref{fig4}, we performed a fine-grained statistical analysis of the action sequences generated by agents in successful tasks. The comparison reveals two distinct interaction patterns:
\begin{itemize}[left=0pt, nosep]
    \item Baseline agents produce execution trajectories dominated by Atomic Actions (e.g., \texttt{\textit{click}}, \texttt{\textit{scroll}}). This indicates that these models are constrained by the need to simulate human mechanical micro-operations, resulting in excessively lengthy interaction sequences.
    \item Eous's execution trajectories are dominated by high-level semantic operations, which effectively replace tedious atomic sequences. This confirms that the CI4A protocol successfully promotes a paradigm shift, specifically an Escalation, from mechanical execution to semantic decision-making. The agent evolves into a tool user capable of recognizing component intent. By invoking deterministic primitives such as \texttt{\textit{setValue}} or \texttt{\textit{NavigateTo }}, Eous achieves an order-of-magnitude collapse in interaction sequences that originally required multi-step physical interactions.
\end{itemize}

\begin{figure}[htbp]
\centering
\includegraphics[width=1\linewidth]{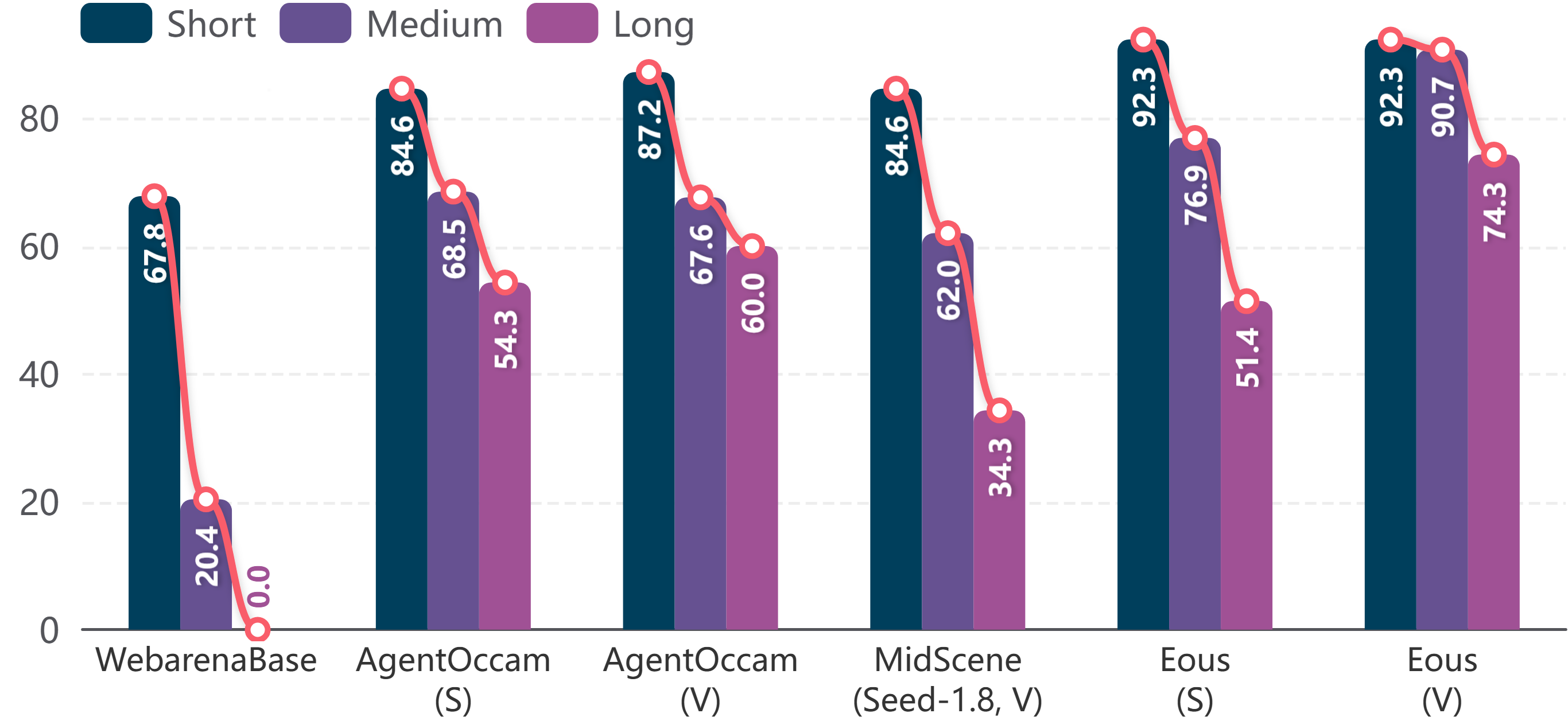}
\caption{\textbf{Performance across task lengths.} Unlike baselines that degrade rapidly in long tasks due to error accumulation, Eous maintains high success rates, demonstrating superior robustness in long-horizon interactions.}
\label{fig5}
\end{figure}

\textbf{$\triangleright$ Semantic shortcuts effectively mitigate error accumulation in long-sequence tasks.} The number of interaction steps in a task typically correlates positively with the probability of error. We categorized the test tasks into three difficulty levels based on the standard number of steps required by the baseline model: Short (1-5 steps), Medium (6-10 steps), and Long ($>$10 steps). We then compared the success rate variations across groups (as shown in Figure \ref{fig5}):
\begin{itemize}[left=0pt, nosep]
    \item Baseline agents exhibit a typical trend of linear decay. As the task sequence lengthens, their average success rates drop precipitously from 81.1\% in the Short to 37.15\% in the Long. This reflects the classic phenomenon of error accumulation—each additional Click-Observe cycle increases the risk of hallucination or positional drift.
    \item Eous demonstrates exceptional performance resilience. In Long group tasks, Eous maintains high success rates of 51.4\% (structure-based) and 74.3\% (vision-based), exhibiting only a marginal decline compared to the Short group. This confirms that CI4A effectively decouples task logic complexity from interaction sequence length. The agent intercepts long-sequence operations, which originally required multiple rounds of trial-and-error exploration, at the decision-making layer, thereby ensuring deterministic performance even in long-horizon, complex tasks.
\end{itemize}

\subsubsection{Case Study}
To analyze how these findings translate to real-world scenarios, we selected two representative complex components Cascader and DatePicker for detailed case studies.

\noindent \textbf{$\triangleright$ Case 1: Cascader.} When interacting with the Cascader component, Eous requires an average of only 1.1 steps, representing a relative reduction of 73.8\% compared to the baseline's 4.2 steps; meanwhile, the accuracy rose from the baseline's 74\% to 100\%. Constrained by the Progressive Disclosure mechanism inherent to the physical interface, baseline agents handling multi-level linkage (e.g., Province / City / District) must execute a lengthy sequence: Click Expand $\rightarrow$ Visual Scan/Scroll $\rightarrow$ Click Target. Crucially, they can only perceive the next level of options after expansion. This blind man feeling an elephant style of interaction incurs extremely high exploration costs. In contrast, benefiting from the Semantic State View $\mathcal{S}$ injected by CI4A, Eous possesses semantic precognition. It acquires all candidate values within the component upon the initial observation without requiring any prerequisite interactions (e.g., \texttt{\{name:"cascader", status:[\{value: "westlake",..\},..]\}}). This enables Eous to bypass the exploratory phase and directly plan the instruction \texttt{call("cascader", "setValue", "westlake"])}, achieving a complexity leap from $O(N)$ to $O(1)$.

\noindent \textbf{$\triangleright$ Case 2: DatePicker.} Eous requires only 1.0 steps on average when interacting with the DatePicker component—a relative reduction of 74.4\% compared to the baseline's 3.9 steps. Baseline agents face rigorous micro-operation challenges during cross-month/year date selection tasks, requiring continuous pagination clicks, grid coordinate identification, and avoidance of disabled date interference. A single click deviation fractures the entire task chain. We observed two typical failure modes: 1) \textit{Format Hallucination}: AgentOccam attempted to bypass the complex date panel by typing text directly into the input field. However, absent clear data constraints, AgentOccam was rejected by frontend validation logic due to its inability to infer the strict input format (e.g., inputting \texttt{\textit{2025/12/31}} when the component required \texttt{\textit{YYYY-MM-DD}}). This format hallucination resulted in invalid operations. 2) \textit{Environmental Fragility}: The MidScene-based pure vision model exhibited extreme sensitivity to the rendering environment. Minor discrepancies in the display configuration of the experimental device caused a systematic drift in screen coordinates, rendering all date selection clicks ineffective during a specific test run. Although this outlier result was discarded, the phenomenon profoundly highlights the robustness defects inherent in pure pixel interaction when facing environmental heterogeneity. In contrast, Eous utilizes the \texttt{\textit{setValue}}) tool to establish a semantic shortcut. By bypassing low-level UI rendering logic and directly injecting a target value that satisfies the type constraints in the interaction metadata (via \texttt{call("datepicker", "setValue", "2025-12-31")})), this mechanism fundamentally circumvents the uncertainty of the physical rendering layer, ensuring absolute precision in execution.

\section{Discussion \& Future Work}
Towards a Symbiotic Web Ecosystem for Humans and Agents. Just as the WAI-ARIA standard resolved accessibility challenges via semantic enhancement, CI4A aims to establish an ``ARIA for Agents'' paradigm. It provides a native, stable, and semantically explicit channel for component-level interaction. We advocate for the adoption of this standard within the frontend ecosystem, enabling Web applications to evolve into environments that are as friendly to agents as they are to humans. Building upon this vision, we discuss the limitations of our current implementation and outline the critical pathways toward this future:

\noindent \textbf{Generalizability of Instrumentation.} Currently, our reference implementation, AntDX, relies on the intrusive instrumentation of the Ant Design framework. While this effectively validates the CI4A protocol, we acknowledge that manually adapting individual UI libraries is labor-intensive. Future work will explore automated instrumentation techniques, leveraging Large Language Models (LLMs) or static analysis to automatically generate CI4A-compliant wrappers for legacy components. This approach aims to reduce the migration costs for existing Web applications.

\noindent \textbf{Scalability to Open-World Scenarios.} Enabling agents to leverage CI4A on websites developed with UI libraries that do not natively support the protocol remains a critical next step.We envision a Skill-based Online Adaptation mechanism. This approach involves identifying generic components within live Web applications and dynamically implementing an ad-hoc CI4A transformation via real-time JavaScript injection. These instantly generated semantic interfaces would be encapsulated as reusable Skills, allowing agents to interact with such websites using the CI4A paradigm and thereby transcending the limitations of closed environments.


\section{Conclusion}
We propose CI4A, a protocol that bridges the execution gap by abstracting UI components into tool primitives. Our CI4A-based agent, Eous, achieves SoTA performance (86.3\%) on WebArena by collapsing physical interaction chains. This work lays the foundation for a transition toward a contract-based, natively friendly Agentic Web.

\bibliographystyle{named}
\bibliography{ijcai26}

\end{document}